\documentclass[letterpaper]{article} 
\usepackage{aaai23}  
\usepackage{times}  
\usepackage{helvet}  
\usepackage{courier}  
\usepackage[hyphens]{url}  
\usepackage{graphicx} 
\usepackage{natbib}  
\usepackage{caption} 
\usepackage{algorithm}
\usepackage{algorithmic}
\usepackage{graphicx}
\usepackage[table]{xcolor}
\usepackage{multirow}
\usepackage{tabularx}
\usepackage{hhline}
\usepackage[para,online,flushleft]{threeparttable}

\definecolor{red}{rgb}{0.95,0.4,0.4}
\definecolor{blue}{rgb}{0.4,0.4,0.95}
\definecolor{darkblue}{rgb}{0,0,0.8}
\definecolor{darkred}{rgb}{0.8,0,0}
\definecolor{darkgreen}{rgb}{0.15,0.6,0.15}
\definecolor{grey}{rgb}{0.6,0.6,0.6}
\definecolor{col1}{RGB}{232, 161, 148}
\definecolor{col2}{RGB}{148, 187, 232}

\definecolor{tgre}{rgb}{0.15,0.6,0.15}
\definecolor{tred}{rgb}{0.6,0.15,0.15}

\usepackage{newfloat}
\usepackage{listings}
\DeclareCaptionStyle{ruled}{labelfont=normalfont,labelsep=colon,strut=off} 
\floatstyle{ruled}
\newfloat{listing}{tb}{lst}{}
\floatname{listing}{Listing}
\nocopyright
\usepackage{algorithm}
\usepackage{algorithmic}
\usepackage{amsthm,amsmath,amssymb}
\usepackage{mathrsfs}
\usepackage{bbm}
\usepackage{amsmath}
\usepackage{multirow}
\usepackage[table]{xcolor}
\usepackage{bbding}
\usepackage{engord}
\usepackage{hyperref}

\title{Crafting Monocular Cues and Velocity Guidance for \\
Self-Supervised Multi-Frame Depth Learning}

\author {
    Xiaofeng Wang\textsuperscript{\rm 1,2},
    Zheng Zhu\textsuperscript{\rm 3},
    Guan Huang\textsuperscript{\rm 3},
    Xu Chi\textsuperscript{\rm 3},
    Yun Ye\textsuperscript{\rm 3},
    Ziwei Chen\textsuperscript{\rm 4},
    Xingang Wang\textsuperscript{\rm 1}
}
\affiliations {
    \textsuperscript{\rm 1} Institute of Automation, Chinese Academy of Sciences\\
    \textsuperscript{\rm 2} School of Artificial Intelligence, University of Chinese Academy of Science\\
    \textsuperscript{\rm 3} PhiGent Robotics 
    \textsuperscript{\rm 4} Southeast University\\
    \{wangxiaofeng2020, xingang.wang\}@ia.ac.cn,  zhengzhu@ieee.org \\
    \{guan.huang, xu.chi, yun.ye\}@phigent.ai, richard\_chen@seu.edu.cn
}

\usepackage{bibentry}

\begin{document}

\maketitle

\begin{abstract}
Self-supervised monocular methods can efficiently learn depth information of weakly textured surfaces or reflective objects. However, the depth accuracy is limited due to the inherent ambiguity in monocular geometric modeling. In contrast, multi-frame depth estimation methods improve the depth accuracy thanks to the success of Multi-View Stereo (MVS), which directly makes use of geometric constraints. Unfortunately, MVS often suffers from texture-less regions, non-Lambertian surfaces, and moving objects, especially in real-world video sequences without known camera motion and depth supervision. Therefore, we propose MOVEDepth, which exploits the \textbf{MO}nocular cues and \textbf{VE}locity guidance to improve multi-frame \textbf{Depth} learning. Unlike existing methods that enforce consistency between MVS depth and monocular depth, MOVEDepth boosts multi-frame depth learning by directly addressing the inherent problems of MVS. The key of our approach is to utilize monocular depth as a geometric priority to construct MVS cost volume, and adjust depth candidates of cost volume under the guidance of predicted camera velocity. We further fuse monocular depth and MVS depth by learning uncertainty in the cost volume, which results in a robust depth estimation against ambiguity in multi-view geometry. Extensive experiments show MOVEDepth achieves state-of-the-art performance: Compared with Monodepth2 and PackNet, our method relatively improves the depth accuracy by 20\% and 19.8\% on the KITTI benchmark. MOVEDepth also generalizes to the more challenging DDAD benchmark, relatively outperforming ManyDepth by 7.2\%. The code is available at \url{https://github.com/JeffWang987/MOVEDepth}.

\end{abstract}

\section{Introduction}
Depth estimation is a fundamental task in 3D computer vision, with versatile applications ranging from virtual/augmented reality \cite{luo2020consistent} to autonomous driving \cite{AndreasGeiger2012AreWR}. Although 3D sensors (e.g., LiDAR, structured light) can generate accurate depth information, it is more attractive to infer depth from a single RGB image in a self-supervised way \cite{RaviGarg2016UnsupervisedCF,ClmentGodard2016UnsupervisedMD,ClmentGodard2018DiggingIS}, which eliminates the necessity of expensive 3D hardware and multi-sensor calibration. However, the accuracy of these monocular methods is not yet on par with 3D sensors due to their inherent ambiguity in geometric modeling.

To improve the monocular depth accuracy, recent multi-frame methods \footnote{Compared with monocular methods that use single frame for inference, the multi-frame methods input $N$ $(N\ge2)$ frames at inference time.} \cite{JamieWatson2021TheTO,FelixWimbauer2020MonoRecSD,RuiWang2019RecurrentNN,HaokuiZhang2019ExploitingTC,ZiyueFeng2022DisentanglingOM,patil2020dont,TaiWang2022Monocular3O} leverage temporal and spatially associations in multi-frame video sequences which are available in real-world scenes (e.g., smart devices \cite{HyowonHa2016HighQualityDF} or moving vehicles \cite{MoritzMenze2015ObjectSF}). Among these approaches,  cost-volume-based methods \cite{ZiyueFeng2022DisentanglingOM,JamieWatson2021TheTO,FelixWimbauer2020MonoRecSD} achieve state-of-the-art depth accuracy as they take advantage of the successful Multi-View Stereo (MVS). However, MVS is still challenged by unsatisfactory reconstructions in real-world scenes with non-Lambertian surfaces, textureless areas, and moving objects \cite{DBLP:journals/tog/KnapitschPZK17,DBLP:conf/cvpr/SchopsSGSSPG17}.
To tackle these problems, teacher-student training architectures \cite{JamieWatson2021TheTO,ZiyueFeng2022DisentanglingOM,DBLP:journals/corr/abs-2205-15034} are proposed to enforce consistency between monocular depth and MVS depth. However, the consistency pushes MVS depth to mimic monocular depth, which underuses the multi-view geometry, thus the performance of these methods is limited.

To improve the multi-view depth accuracy, the learning-based MVS methods \cite{DBLP:conf/eccv/YaoLLFQ18,DBLP:conf/cvpr/0008LLSFQ19} densely sample depth candidates in a large range. However, the dense sampling strategy causes matching ambiguity in real-world video frames without known camera pose and depth supervision (see Fig. \ref{fig:kitti_vis}). To mitigate the problem, we explore an efficient approach to enhance geometric cues for improving self-supervised multi-frame depth learning. Our intuition is that the monocular depth serves as a geometric priority of the scene, and the multi-frame matching ambiguity can be significantly reduced by sampling depth candidates near the monocular priority. Apart from the matching ambiguity, the multi-view geometry is still challenged by insufficient \textit{Triangulation Prior} \cite{JohannesLSchonberger2016PixelwiseVS},  especially in static/slow video sequences where nearby frames share little stereo baseline. To address the problem, the predicted camera velocity is leveraged to adaptively adjust the depth range. Specifically, multi-frames with higher motion velocity have larger viewpoint change, which can benefit multi-view geometry, so the depth range is enlarged to infer more accurate depth. In contrast, static frames carry little information for depth inference, thus the depth range is shrunk to the more reliable monocular priority. Besides, we fuse monocular depth and MVS depth by learning uncertainty in the cost volume, resulting in a robust depth estimation against artifacts in multi-view geometry (e.g., moving objects, textureless areas).

Owing to the monocular depth priority and camera velocity guidance, MOVEDepth achieves state-of-the-art performance: Compared with competitive monocular baselines \cite{ClmentGodard2018DiggingIS,VitorGuizilini20193DPF}, our method relatively improves the depth accuracy by $\sim$20\% on the KITTI benchmark. MOVEDepth also generalizes to the more challenging DDAD benchmark, relatively outperforming ManyDepth \cite{JamieWatson2021TheTO} by 7.2\%. Besides,  qualitative analysis demonstrates that our method is more robust against challenging artifacts where multi-view geometry fails.

The main contributions are three-fold as follows:

- We propose a novel self-supervised multi-frame depth learning framework, named MOVEDepth. It leverages the monocular depth cues as a geometric priority, and the multi-frame matching ambiguity is mitigated by sampling depth candidates near the monocular priority.

- The velocity-guided depth sampling is proposed to address failure cases caused by slow/static camera motion. And an adaptive fusing layer is introduced to learn uncertainty in cost volume, which mitigates artifacts brought by textureless areas and moving objects.

- We conduct extensive experiments on KITTI and DDAD, and the results show our method achieves superior depth accuracy in the complex real-word scenes with fewer depth candidates.


\begin{figure*}[ht]
  \centering
  \includegraphics[width=1\textwidth]{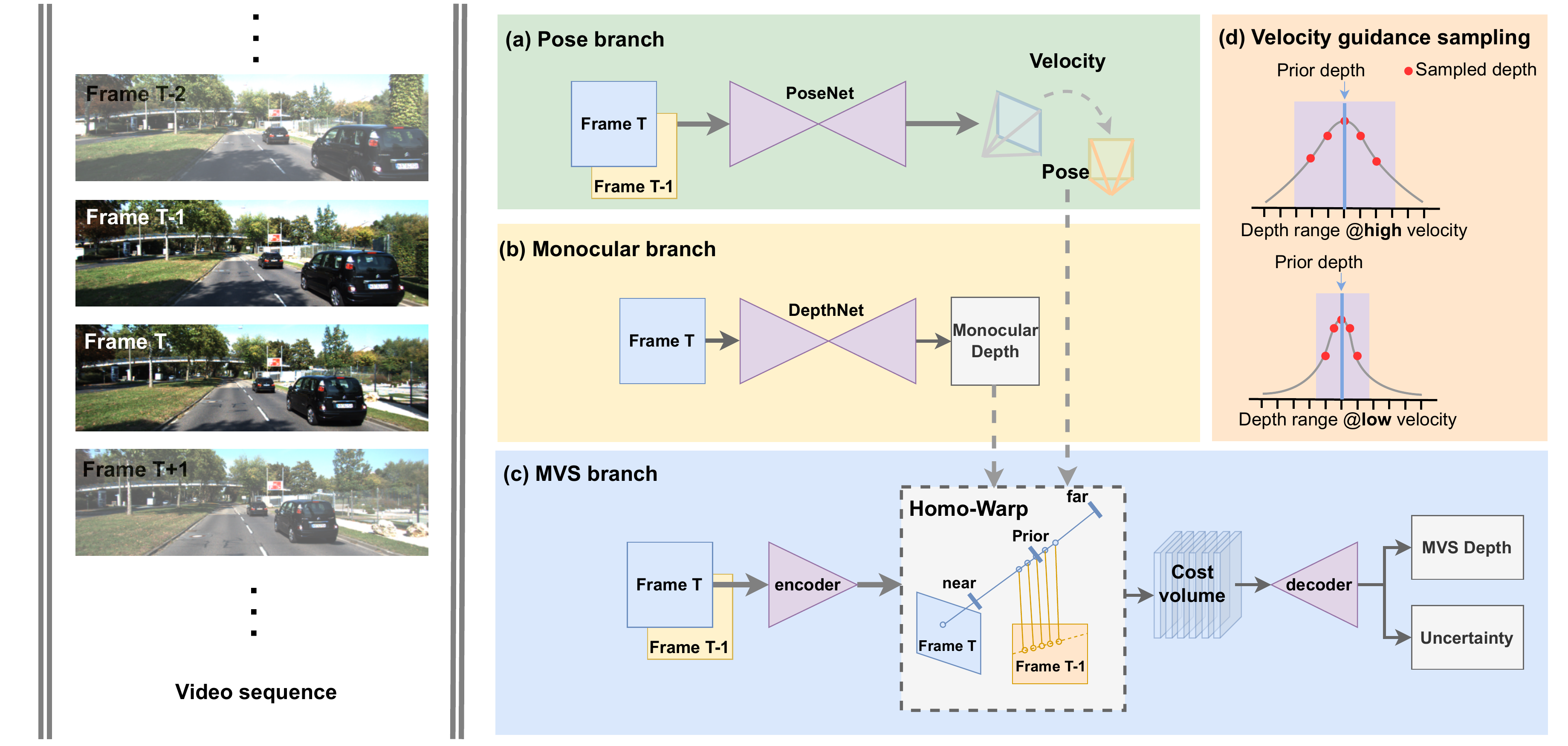}
  \caption{The main network architecture of MOVEDepth.
  \textbf{(a)} PoseNet is utilized to estimate camera ego-motion and velocity between frame $T$ and frame $T-1$. \textbf{(b)} The monocular depth is predicted using the DepthNet, which serves as a geometric priority to construct MVS cost volume. \textbf{(c)} We conduct homography warping between the encoded frame features using the predicted camera ego-motion and monocular depth priority. The resulting cost volume is decoded into a depth map and an uncertainty map. \textbf{(d)} The depth sampling range of homography warping is adaptively adjusted under the guidance of predicted camera velocity.}
  \label{fig:m2d}
  \end{figure*}

\section{Related Work}
In this section, we review depth estimation approaches relevant to our method in the following two categories: monocular depth learning and multi-frame depth learning.
\subsection{Monocular Depth Learning} 
Despite the inherently ill-posed problems of geometric reasoning, monocular depth learning has been studied extensively in the literature. Supervised methods exploit sparse annotations \cite{WeifengChen2016SingleImageDP} or dense points clouds from 3D sensors \cite{DavidEigen2014DepthMP,DavidEigen2014PredictingDS,HuanFu2018DeepOR}. Self-supervised approaches mitigate the expensive 3D sensors and human annotations, instead training with photo-metric consistency losses using stereo images \cite{JunyuanXie2016Deep3DFA,RaviGarg2016UnsupervisedCF}. Recently, monocular video supervision \cite{TinghuiZhou2017UnsupervisedLO,ClmentGodard2018DiggingIS,VitorGuizilini20193DPF} is an attractive alternative to stereo-based methods but it brings a new set of challenges: The model needs to jointly estimate camera ego-motion and image depth to form reprojection losses, which further increases the difficulty of monocular geometric modeling. Although there still exists a huge performance gap between monocular methods and other geometric methods, recent advances in monocular depth learning provide a plausible geometric priority, which can benefit other geometric approaches to refine the depth.

\subsection{Multi-Frame Depth Learning}
Given multiple 2D images and camera parameters, MVS can reconstruct the dense geometry of the scene \cite{JohannesLSchonberger2016PixelwiseVS}. Recent advances in learning-based MVS approaches \cite{DBLP:conf/eccv/YaoLLFQ18,DBLP:conf/cvpr/GuFZDTT20,wang2022mvster} further improve the reconstruction accuracy. However, these methods assume known camera poses and static scenes. MonoRec \cite{FelixWimbauer2020MonoRecSD} extends MVS in dynamic environments that are captured with a single moving camera, but it needs the pre-trained segmentation network \cite{DBLP:conf/iccv/HeGDG17} and sparse depth obtained by a visual odometry system \cite{NanYang2018DeepVS}. MaGNet \cite{Bae2022} fuses single-view probability with multi-view geometry to produce robust estimations. However, it still requires ground truth depth supervision. The self-supervised method \cite{JamieWatson2021TheTO} removes the requirement for human annotations and depth supervision, and it proposes a teacher-student training architecture to encourage the network to ignore unreliable regions in MVS cost volume. \cite{ZiyueFeng2022DisentanglingOM} further improves the depth accuracy in regions of dynamic objects by disentangling object motions, and \cite{DBLP:journals/corr/abs-2205-15034} utilizes Deformable Convolution Networks (DCNs \cite{DBLP:conf/iccv/DaiQXLZHW17}) to enhance the depth estimates in low-texture and homogeneous-texture regions. However, these methods enforce consistency between MVS depth and monocular depth, which underuses the geometric reasoning of MVS. In contrast, the proposed MOVEDepth directly addresses the inherent problems of MVS (e.g., static/slow camera motion, object motions, and textureless regions).
  
\section{Method}
A detailed description of MOVEDepth is given in this section and the main network architecture is illustrated in Fig.~\ref{fig:m2d}. Given a video sequence, (a) we firstly utilize PoseNet to estimate camera ego-motion and velocity between frame $T$ and frame $T-1$. (b) Then the monocular depth is predicted using the DepthNet. (c) Subsequently, we conduct homography warping between the encoded frame features using the predicted camera ego-motion and monocular depth priority. The resulting cost volume is decoded into a depth map and an uncertainty map, which serves as guidance for fusing the monocular depth and MVS depth. (d) Significantly, the depth sampling range of homography warping is adaptively adjusted under the guidance of predicted camera velocity, which mitigates problems brought by slow/static camera motion.

The following parts start with the preliminary of self-supervised monocular depth learning from a video sequence. We then introduce the important innovations to improve multi-frame depth learning by fusing monocular cues.
\subsection{Self-Supervised Monocular Depth Learning}
 The self-supervised pipeline is conducted by jointly training a DepthNet $\theta_{\text{d}}$ (see Fig.~\ref{fig:m2d}(b)) and a PoseNet $\theta_{\text{p}}$ (see Fig.~\ref{fig:m2d}(a)) \cite{TinghuiZhou2017UnsupervisedLO}, and they are trained only on video frames $\{\mathbf{I}_t\}_{t=1}^N$. Specifically, we estimate the monocular depth $D_\text{Mono}=\theta_{\text{d}}(\mathbf{I}_t)$ of current frame $\mathbf{I}_t$, and predict relative camera pose $\mathbf{\left[\mathbf{R} \mid \mathbf{T}\right]}_{t\rightarrow t+k}=\theta_{\text{p}}(\mathbf{I}_t, \mathbf{I}_{t+k})$ between frame $\mathbf{I}_t$ and frame $\mathbf{I}_{t+k}(k\in{-1,1})$. Then, we can synthesize $\mathbf{I}_t$ from viewpoint $\mathbf{I}_{t+k}$ by the following operation:
\begin{equation}
    \mathbf{I}_{t+k \rightarrow t}(D_{\text{Mono}})=\mathbf{I}_{t+k}\left\langle\operatorname{proj}\left(D_\text{Mono}, \mathbf{\left[\mathbf{R} \mid \mathbf{T}\right]}_{t \rightarrow t+k}, \mathbf{K}\right)\right\rangle,
\end{equation}
where $\mathbf{K}$ is the camera intrinsics, $\operatorname{proj}(\cdot)$ is the projection function that returns the 2D pixel coordinates of the projected $D_\text{Mono}$, and $\left\langle\cdot\right\rangle$ is the pixel sampling operator. Following the optimization convention \cite{ClmentGodard2018DiggingIS}, the training pipeline is optimized by a reprojection loss:
\begin{equation}
    \mathcal{L}_\text{r}(D_{\text{Mono}})=\min_{k} p e\left(\mathbf{I}_{t}, \mathbf{I}_{t+k \rightarrow t}(D_{\text{Mono}})\right),
    \label{eq:reprj}
\end{equation}
where the $\min$ operation selects the best matching frames to avoid ambiguity brought by occlusions, and $pe(\cdot)$ is a weighted combination of $\mathcal{L}_1$ loss and structure similarity (SSIM) loss. The reprojection loss is calculated over multi-scale depth outputs, and more implementation details can be found in \cite{ClmentGodard2018DiggingIS}.

\subsection{MOVEDepth Design}
\subsubsection{MVS Depth from Monocular Depth Priority} Multi-view approaches warp source images into the reference camera frustum to form cost volume, and estimate depth to be the highest-activated value in cost volume \cite{DBLP:conf/eccv/YaoLLFQ18}. Although the hard-coded multi-view methods reduce geometry ambiguity and generate more accurate depth, they are still challenged by texture-less regions, non-Lambertian surfaces, and moving objects, especially in real-world video frames without known camera motion. The monocular methods, on the other hand, are more robust against weakly textured regions or moving objects but the overall depth accuracy is limited. Therefore, we exploit monocular cues to complement the limitation of MVS (see Fig.~\ref{fig:m2d}(c)), which is elaborated in the following.

Given a current frame $\mathbf{I}_{t}\in \mathbb{R}^{H \times W \times 3}$ and its nearby frame $\mathbf{I}_{t-1} \in \mathbb{R}^{H \times W \times 3}$ (the future frames is not used to enable online depth prediction), we firstly leverage a encoder $\theta_{\text{enc}}$ to extract 2D features of these frames, where the images are downscaled to lower resolution deep features $\mathbf{F}_{i(i\in\{0,-1\})}\in \mathbb{R}^{H/4 \times W/4 \times C}$. Following previous learning-based MVS \cite{DBLP:conf/eccv/YaoLLFQ18,DBLP:conf/cvpr/GuFZDTT20}, the plane sweep stereo \cite{RobertTCollins1996ASA} is utilized to establish multiple front-to-parallel planes in the current frame. Specifically, equipped with camera intrinsic $\mathbf{K}$ and extrinsic $\left[\mathbf{R} \mid \mathbf{T}\right]$ estimated by PoseNet $\theta_{\text{p}}$, the previous frame features can be warped into the current camera frustum:

\begin{equation}
\label{eq:homo}
    \mathbf{p}_{t-1,j}=\mathbf{K} \cdot\left(\mathbf{R} \cdot\left({\mathbf{K}}^{-1} \cdot \mathbf{p}_{t} \cdot d_{j}\right)+\mathbf{T}\right),
\end{equation}
where $d_j$ is the $j$-th hypothesized depth candidates of pixel $\mathbf{p}_t$ in the current frame feature $\mathbf{F}_t$, and $\mathbf{p}_{t-1,j}$ denotes the corresponding pixel in the previous frame feature $\mathbf{F}_{t-1}$. After the warping operation, the volume feature $\mathbf{V}_{t-1}\in\mathbb{R}^{H/4 \times W/4 \times C \times D}$ is constructed, where $D$ is the number of depth candidates. 
Significantly, to reduce depth searching space, we specify the depth range $\mathcal{R}$ using the  monocular depth priority $D_\text{Mono}$:
\begin{equation}
    \mathcal{R} = \{d|d_\text{min}\le d\le d_\text{max}\},
\end{equation}
where $(d_\text{min}+d_\text{max})/2=D_\text{Mono}$, and $d_\text{min}, d_\text{max}$ is adaptively adjusted under the guidance of camera velocity, which is elaborated in the next subsection.

Given previous frame volume $\mathbf{V}_{t-1}$, we then use \textit{group correlation} \cite{DBLP:conf/cvpr/WangGVSP21,QingshanXu2019LearningID} to construct cost volume, which measures the visual similarity between the current frame and the previous frame:

\begin{equation}
    \mathbf{s}_i^{g}=\frac{1}{G}\left\langle \mathbf{v}_i^g,\mathbf{f}_i^g\right\rangle,
\end{equation}
where $\mathbf{v}_i^g\in \mathbb{R}^{\frac{C}{G}\times D}$ is the $g$-th group feature of $\mathbf{v}_i$ ($\mathbf{v}_i\in \mathbb{R}^{C\times D}$ is the $i$-th pixel feature of $\mathbf{V_t}$), and $\mathbf{f}_i^g\in \mathbb{R}^{ \frac{C}{G}\times 1}$ is the $g$-th group feature of $\mathbf{f}_i$ ($\mathbf{f}_i$ is the $i$-th pixel feature of $\mathbf{F}_t$), and $\left\langle \cdot,\cdot\right\rangle$ is the inner product. Then $\left\{\mathbf{s}_i^g\right\}_{g=0}^{G-1}$ are channel-wise stacked to generate $\mathbf{s}_i\in \mathbb{R}^{G \times D}$, which is $i$-th pixel feature of the final cost volume $\mathbf{S}_t\in\mathbb{R}^{H/4\times W/4\times G \times D}$.

The calculated cost volume is subsequently decoded by a light-weight $\theta_\text{dec}$ to get depth probability $\mathbf{P}\in\mathbb{R}^{H/4\times W/4 \times D}$, and the MVS depth is generated by \textit{localmax} \cite{DBLP:journals/corr/abs-2112-05126}:
\begin{equation}
    D_\text{MVS}(\mathbf{p})=\left(\frac{1}{\sum_{j=\mathbf{X}(\mathbf{p})-r}^{\mathbf{X}(\mathbf{p})+r} \mathbf{p}_j} \sum_{j=\mathbf{X}(\mathbf{p})-r}^{\mathbf{X}(\mathbf{p})+r} \frac{1}{d_{j}} \cdot \mathbf{p}_j\right)^{-1},
\label{eq:localmax}
\end{equation}
where $\mathbf{p} \in \mathbb{R}^{D}$ is the pixel value of $\mathbf{P}$, $\mathbf{X}(\mathbf{p})=\text{argmax}_j \mathbf{p}_j$ is the index of the highest value for $\mathbf{p}$, and $r$ is a radius parameter (typically set as 1). Finally, the \textit{convex interpolation} \cite{ZacharyTeed2022RAFTRA} is leveraged to upsample the MVS depth to the original resolution.

\subsubsection{Velocity-Guided Depth Sampling}
In the previous subsection, the monocular depth is leveraged as a geometric center for depth sampling, but the depth range is left to be addressed. Typically,
learning-based MVS \cite{DBLP:conf/eccv/YaoLLFQ18,DBLP:conf/cvpr/0008LLSFQ19} sample depth candidates in a fixed range, which is either calculated by COLMAP \cite{JohannesLSchonberger2016PixelwiseVS} or learned by networks \cite{JamieWatson2021TheTO}. However, the depth range is utilized to describe the entire scene, and densely searching in such a wide range is computationally expensive and can not produce accurate depth \cite{DBLP:conf/cvpr/GuFZDTT20}. Recent methods reduce depth range by coarse-to-fine sampling \cite{DBLP:conf/cvpr/GuFZDTT20,DBLP:conf/cvpr/WangGVSP21} or confidence-based sampling \cite{Bae2022,DBLP:conf/cvpr/ChengXZLLRS20}. However, we empirically find these sampling strategies are limited in self-supervised multi-frame depth learning (see Tab. \ref{tab:abla_vel}), as they overlook the \textit{Triangulation Prior} \cite{JohannesLSchonberger2016PixelwiseVS} of nearby frames.

To mitigate the problem, we propose velocity-guided depth sampling. The key innovation is to associate \textit{Triangulation Prior} with camera motion velocity $v$. Namely, the viewpoint changes noticeably when the camera moves at a high velocity, providing a sufficient \textit{Triangulation Prior} for multi-view geometry. In contrast, slow/static video frames share a similar viewpoint, thus the \textit{Triangulation Prior} is limited (theoretical analysis is in supplement). For video frames with sufficient \textit{Triangulation Prior}, we expand depth range to infer accurate depth, and for frames with insufficient \textit{Triangulation Prior}, the depth range is shrunk to the more reliable monocular priority. The depth sampling range is specified as follows:

\begin{equation}
\begin{aligned}
    d_\text{min}&=D_\text{Mono}(1-\beta\mathcal{T}(v))\\ 
    d_\text{max}&= D_\text{Mono}(1+\beta\mathcal{T}(v)),
\end{aligned}
\end{equation}
 where the camera motion velocity $v=\alpha \|\mathbf{T}\|_2$ is the byproduct of PoseNet $\theta_\text{p}$  ( $\mathbf{T}$ is the camera translation estimated by $\theta_\text{p}$, and $\alpha$ is the camera frame rate).  $\beta$ is a hyper-parameter, and $\mathcal{T}(\cdot)$ is a scale function that transforms $v$ to a real-world scale, which can be calculated by median-scaling \cite{ClmentGodard2018DiggingIS} or camera-height-scaling \cite{DBLP:conf/iccv/YinWDC17}. To ensure training stability$, \beta\mathcal{T}(\cdot)$ is clamped to range (0, 1). 
 
 Notably, the depth sampling strategy resembles \textit{Gaussian Sampling} with mean of $D_\text{Mono}$ and variance of $\beta\mathcal{T}(v)$, and the MVS depth range shrinks to the more reliable monocular depth when the camera is static. Differently, the depth candidates are not sampled by their probability, but by a deterministic inverse sampling strategy:
\begin{equation}
    d_{j}=\left(\left(\frac{1}{d_{\min }}-\frac{1}{d_{\max }}\right) \frac{j}{D-1}+\frac{1}{d_{\max }}\right)^{-1},
\end{equation}
where $j=0 \ldots D-1$. Compared with linear sampling \cite{DBLP:conf/eccv/YaoLLFQ18} or probabilistic sampling \cite{Bae2022}, inverse depth sampling results in uniformly distributed depth candidates at the pixel level, which is beneficial for large-scale multi-frame matching \cite{QingshanXu2019LearningID}.
\subsubsection{Uncertainty-Based Depth Fusing}
 \begin{figure}[t]
  \centering
  \includegraphics[width=1\columnwidth]{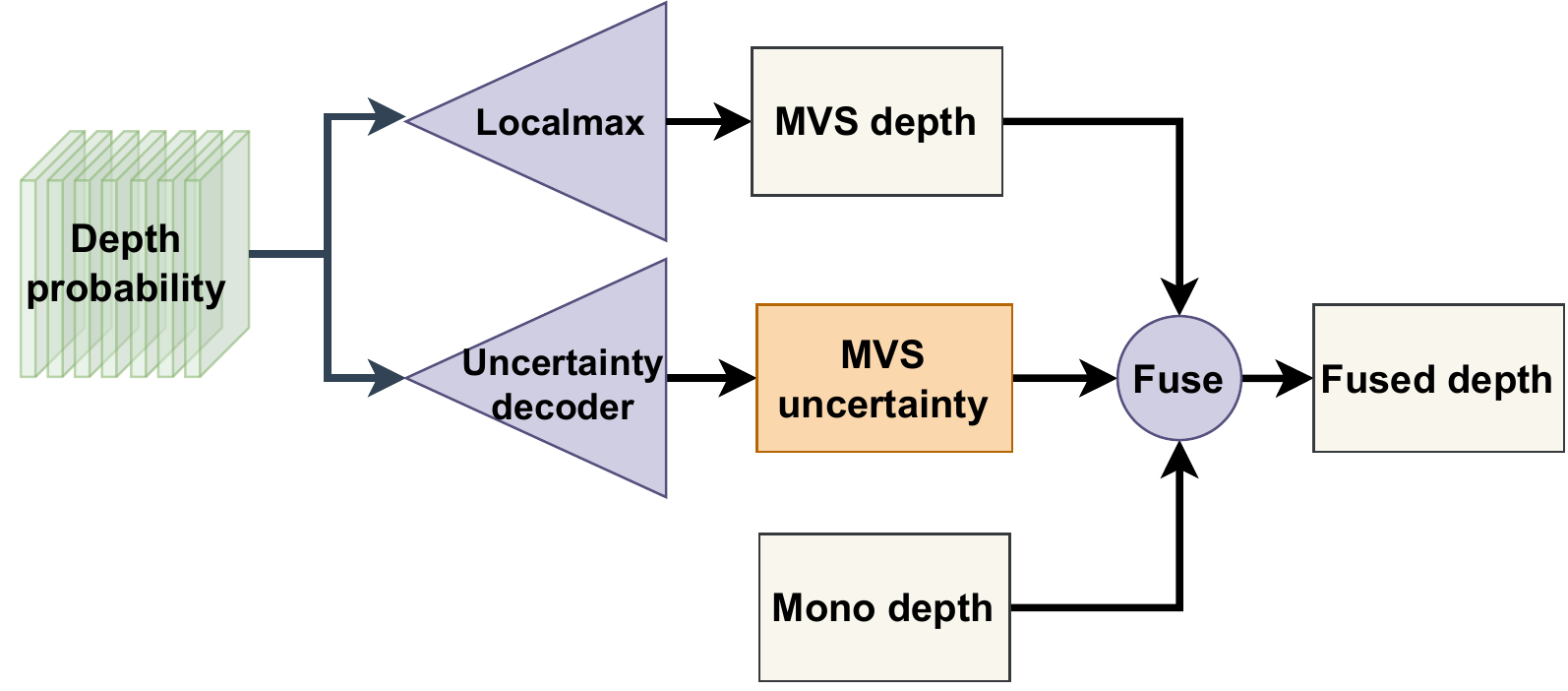} 
  \caption{MOVEDepth learns uncertainty in depth probability to fuse monocualr depth and MVS depth. The upper branch decodes depth probability into depth by \textit{localmax} (Eq. \ref{eq:localmax}), and the lower branch decodes depth probability into an uncertainty map, which serves as a guidance for fusing MVS depth and monocular depth.}
  \label{fig:fuse}
  \end{figure}
  
The calculated $D_\text{MVS}$ is still challenged by texture-less regions, non-Lambertian surfaces, and moving objects, which are the inherent problems of the multi-view geometry. To alleviate the problem, the uncertainty-based fusing method is introduced to replace unsatisfactory $D_\text{MVS}$ with the more reliable $D_\text{Mono}$. As shown in Fig.~\ref{fig:fuse}, we leverage an \textit{Uncertainty Decoder} $\theta_\text{u}$ to learn uncertainty map $\mathbf{U}$ from the entropy of depth probability $\mathbf{p}$:
\begin{equation}
    \mathbf{U}(\mathbf{p}) = \theta_\text{u}(\sum_{j=0}^{D-1}-\mathbf{p}_{j} \log \mathbf{p}_{j}),
\end{equation}

\newcommand{\x}{ $\times$ }
\newcommand{\midline}{  }

\newcommand{\splitline}{\arrayrulecolor{black}\hhline{~------------}}

\newcommand*\rot{\rotatebox{90}}

\definecolor{Asectioncolor}{RGB}{255, 200, 200}
\definecolor{Bsectioncolor}{RGB}{255, 228, 196}
\definecolor{Csectioncolor}{RGB}{235, 255, 235}
\definecolor{Dsectioncolor}{RGB}{235, 235, 255}

\begin{table*}[ht]
  \centering
  \small
  \resizebox{1.0\textwidth}{!}{
    \begin{tabular}{l|c|c|c|c|c|c|c|c|c}
        \arrayrulecolor{black}\hline
          Method & Test frames  & W\x H & \cellcolor{col1}Abs. Rel. & \cellcolor{col1}Sq. Rel. & \cellcolor{col1}RMSE  & \cellcolor{col1}$\text{RMSE}_{\text{log}}$ & \cellcolor{col2}$\delta < 1.25 $ & \cellcolor{col2}$\delta < 1.25^{2}$ & \cellcolor{col2}$\delta < 1.25^{3}$ \\
         
        \hline\hline
      \noalign{\smallskip}
         Ranjan \cite{ranjan2018adversarial}  & 1 & 832\x 256 & 0.148 & 1.149 & 5.464 & 0.226 & 0.815 & 0.935 & 0.973\\
        \midline
         EPC++ \cite{luo2019every} & 1  & 832\x256 & 0.141 & 1.029 & 5.350 & 0.216 & 0.816 & 0.941 & 0.976\\
        \midline
          Struct2depth (M) \cite{casser2018depth}  & 1  &416\x128& 0.141 & 1.026 & 5.291 &  0.215 & 0.816 & 0.945 & 0.979\\
        \midline
          Videos in the wild \cite{gordon2019depth} & 1  &  416\x128& 
        0.128 & 0.959  & 5.230 & 0.212 & 0.845 & 0.947 & 0.976 \\
        
        \midline
          Guizilini  \cite{guizilini2020semantically} & 1 & 640\x192 & 0.102 & 0.698 & 4.381& 0.178 & 0.896 & 0.964 & 0.984 \\
        \midline
         Johnston ~\cite{johnston2020self} & 1 & 640\x192 & 0.106 & 0.861 & 4.699 & 0.185 & 0.889 & 0.962 & 0.982 \\
        \midline
          Monodepth2 \cite{ClmentGodard2018DiggingIS} & 1  &  640\x192 &
         {0.115} &   {0.903} &   {4.863} &   {0.193} &   {0.877} &   {0.959} &   {0.981} \\ 
         \midline
         PackNet-SFM \cite{VitorGuizilini20193DPF} & 1 & 640\x 192 & 0.111 & 0.785 & 4.601 & 0.189 & 0.878 & 0.960 & 0.982 \\
        \midline
         Li \cite{li2020unsupervised}  & 1    &  416\x128 &
        0.130 & 0.950 & 5.138 & 0.209 & 0.843 & 0.948 & 0.978 \\
        \midline
        R-MSFM \cite{ZhongkaiZhou2022RMSFMRM}& 1 & 1024\x320 & 0.108 & 0.748 & 4.481 & 0.179 & 0.893 & 0.963 & 0.982 \\
        \midline
        RM-Depth \cite{TakWaiHui2022RMDepthUL}& 1 & 640\x192 & 0.108 & 0.710 & 4.513 & 0.183 & 0.884 & 0.964 & 0.983 \\
        
        \noalign{\smallskip}
        \hline
        \noalign{\smallskip}
         GLNet \cite{chen2019self} & 3 (-1,  0, +1)    & 416\x128  & 0.099  & 0.796  & 4.743  & 0.186  & 0.884  & 0.955  & 0.979 \\
        \midline
         Luo  \cite{luo2020consistent} & N & 384\x112 &  0.130 &  2.086 & 4.876 & 0.205 & 0.878 & 0.946 &  0.970\\
        \midline
         CoMoDA \cite{kuznietsov2021comoda} &  N &   640\x192 & 0.103 & 0.862 & 4.594 &  0.183  & 0.899 &  0.961 &  0.981 \\
         \midline
          Patil \cite{patil2020dont}  & N    &  640\x192  & 0.111  & 0.821  & 4.650  & 0.187  & 0.883  & 0.961  & 0.982 \\
        \midline
         TC-Depth \cite{DBLP:conf/3dim/RuhkampGCNB21} &  3(-1, 0, +1)   &  640\x192 & 0.103 & 0.746 & 4.483 &  0.185  & 0.894 &  - &  0.983 \\
        ManyDepth \cite{JamieWatson2021TheTO}  & 2 (-1, 0)  & 640\x192 &   0.098  &   0.770  &   4.459  &   0.176  &   0.900  &   0.965  &   0.983 \\
        \midline
         DynamicDepth \cite{ZiyueFeng2022DisentanglingOM}  & 2 (-1, 0)  & 640\x192 &   0.096  &   0.720  &   4.458  &   0.175  &   0.897  &   0.964  &   0.984 \\
         
        \noalign{\smallskip}
        \hline
        \noalign{\smallskip}
        
         $\text{MOVEDepth}$ (with Monodepth2) & 2 (-1, 0) &  640\x192 &   0.092  &   0.686  &   4.332  &   0.173  &   \textbf{0.904}  &   \textbf{0.966}  &   0.983  \\
         \midline
         $\textbf{MOVEDepth (with PackNet)}$ & 2 (-1, 0) &  640\x192 &   \textbf{0.089}  &   \textbf{0.663}  &   \textbf{4.216}  &   \textbf{0.169}  &   \textbf{0.904}  &   \textbf{0.966}  &   \textbf{0.984}  \\

         \noalign{\smallskip}
        \arrayrulecolor{black}\hline

    \end{tabular}
  } 
  \caption{Comparison of $\text{MOVEDepth}$ to existing self-supervised methods on the KITTI~\cite{AndreasGeiger2012AreWR} Eigen split. At top we compare $\text{MOVEDepth}$ with monocular methods using one frame at test time. At middle we compare  $\text{MOVEDepth}$ with multi-frame methods with multiple frames inputs at test time. At bottom we show our method with different monocular depth priority.} 
    \label{tab:kitti_eigen} 
\end{table*}

where $\theta_\text{u}$  is comprised of 2D Convolutional Neural Network (CNN) blocks and a \textit{Sigmoid} function.  The reason for adopting the entropy is that the randomness of the depth probability distribution is positively related to the MVS depth uncertainty \cite{DBLP:conf/bmvc/ZhangYLLF20}. Subsequently, the uncertainty map is leveraged to calculate the fused depth $D_\text{Fuse}$:
\begin{equation}
    D_\text{Fuse} = \mathbf{U}\odot D_\text{Mono} + (\mathbf{1}-\mathbf{U}) \odot D_\text{MVS},
\end{equation}
where $\odot$ denotes element-wise product.

\subsubsection{Loss Function}
MOVEDepth is end-to-end trained in a self-supervised manner, and the loss consists of three parts:
\begin{equation}
 \mathcal{L}_{\text{MOVEDepth}} = \lambda_1 \mathcal{L}(D_\text{Mono}) +
 \lambda_2 \mathcal{L}(D_\text{MVS}) + \lambda_3 \mathcal{L}(D_\text{Fuse}),
\end{equation}
where $\lambda_1,\lambda_2,\lambda_3$ are the loss weights, and $\mathcal{L}(\cdot)$ is a weighted combination of reprojection loss $\mathcal{L}_\text{r}$ (Eq. (\ref{eq:reprj})) and depth smooth loss $\mathcal{L}_\text{s}$ \cite{ClmentGodard2016UnsupervisedMD}:
\begin{equation}
    \mathcal{L}(D)=\mathcal{L}_\text{r}(D)+\gamma \mathcal{L}_\text{s}(D),
\end{equation}
where $\gamma$ denotes the loss weight.

\section{Experiment}
\subsection{Datasets}
MOVEDepth is evaluated on KITTI \cite{AndreasGeiger2012AreWR} and DDAD \cite{VitorGuizilini20193DPF}  to verify the effectiveness. KITTI is an outdoor dataset in the driving scenario, which is the standard benchmark for depth evaluation. Following from the Eigen split \cite{DavidEigen2014PredictingDS}, with data preprocessing from \cite{TinghuiZhou2017UnsupervisedLO}, the data is divided into 39810/4424/697 training, validation and test images.
For DDAD, it is a diverse dataset of the highway, urban, and residential scenes curated by self-driving cars, which is a novel benchmark for depth evaluation. Notably, DDAD evaluates with longer depth ranges and denser LiDAR ground-truth, which is particularly challenging for multi-frame methods. Following \cite{VitorGuizilini20193DPF}, only front-view images are used, resulting in 12560/3950 training and validation images.

\subsection{Implementation Details}
Following previous work \cite{ClmentGodard2018DiggingIS,JamieWatson2021TheTO}, we use color-jitter and flip as training-time augmentations, and MOVEDepth is trained with an input resolution of 640$\times$192 (KITTI) and 640$\times$384 (DDAD). We only use two frames $\{\mathbf{I}_{t-1},\mathbf{I}_{t}\}$ for cost volume construction, and use $\{\mathbf{I}_{t-1},\mathbf{I}_{t},\mathbf{I}_{t+1}\}$ for reprojection loss. 
We train MOVEDepth for 20 epochs and optimize it with Adam \cite{DBLP:journals/corr/KingmaB14}. MOVEDepth is trained on 4 NVIDIA RTX 3090 GPUs with batch size 6 on each GPU. The learning rate is initially set as 0.0002, which decays by a factor of 10 for the final 5 epochs. The feature extractor in $\theta_\text{d}$ and $\theta_\text{p}$ follows the same architecture in \cite{ClmentGodard2018DiggingIS,VitorGuizilini20193DPF}, and $\theta_\text{enc}$, $\theta_\text{dec}$ comprise light-weight 2D CNNs and 3D CNNs (more network details are in supplement). Following \cite{ClmentGodard2018DiggingIS,JamieWatson2021TheTO}, the loss weight $\gamma$ is set as 0.001, and $\lambda_{i(i\in \{1,2,3\})}=1$. For MVS cost volume construction, the number of depth candidates is 16, group correlation $G=16$, and $\beta=0.15$.

\begin{figure*}[ht]
  \centering
  \includegraphics[width=1\textwidth]{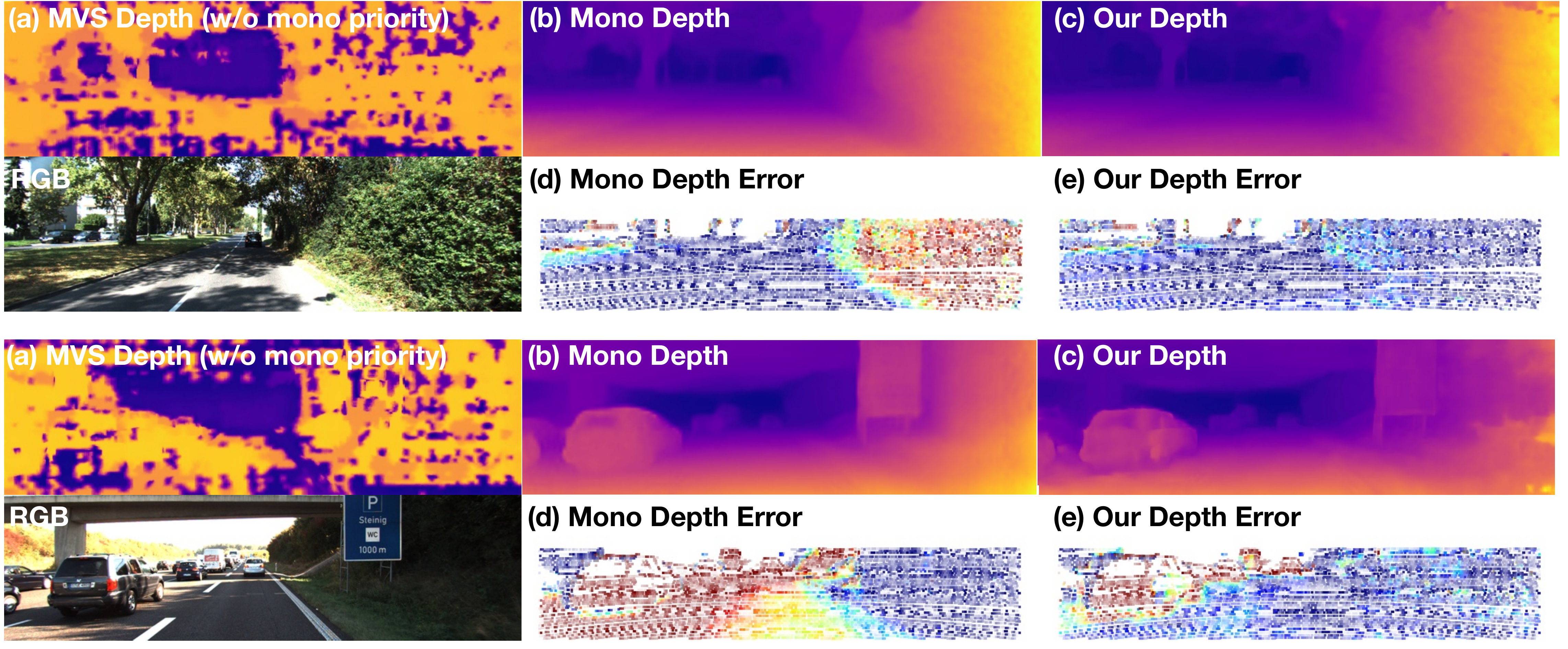}
  \caption{Qualitative results on  KITTI. We visualize depth prediction from \textbf{(a)} MVS baseline (MOVEDepth without monocular priority), \textbf{(b)} monocular method \cite{ClmentGodard2018DiggingIS}, and \textbf{(c)} MOVEDepth. \textbf{(d)(e)} We also visualize the absolute relative error (Abs. Rel.) compared to the ground truth, where the error ranges from blue (Abs. Rel. = 0.0) to red (Abs. Rel. = 0.2). }
  \label{fig:kitti_vis}
  \end{figure*}
  
\subsection{KITTI Results}
We compare MOVEDepth with monocular methods and multi-frame methods, and the quantitative results are shown in Tab.~\ref{tab:kitti_eigen}, where MOVEDepth achieves a state-of-the-art performance among all the competitors. We observe that MOVEDepth relatively improves the best-performing monocular method \cite{guizilini2020semantically} by 12.7\% (on Abs. Rel.), which demonstrates it is significant to investigate multi-frame geometry to improve depth accuracy. Besides, MOVEDepth relatively outperforms multi-frame methods (ManyDepth \cite{JamieWatson2021TheTO} and DynamicDepth \cite{ZiyueFeng2022DisentanglingOM}) by 9.2\% and 7.3\% (on Abs. Rel.), which verifies that our method is effective for multi-frame depth learning. Furthermore, we implement MOVEDepth with depth priority from different monocular methods, and our method relatively improves Monodepth2 \cite{ClmentGodard2018DiggingIS} and PackNet \cite{VitorGuizilini20193DPF} by 20\% and 19.8\% (on Abs. Rel.), which verifies that MOVEDepth  benefits from advances in existing monocular methods. Notably, the resolution of depth priority is $\frac{1}{4}$ of the inputs, and the depth priority significantly reduces the depth candidates of MVS, thus MOVEDepth (with Monodepth2) still runs at an acceptable speed (32 FPS) at single RTX 3090.

Qualitative results are presented in Fig.~\ref{fig:kitti_vis}, where the MVS baseline (without monocular priority) fails in video-based depth learning due to matching ambiguity in a large depth range. In contrast, monocular method \cite{ClmentGodard2018DiggingIS} and MOVEDepth can generate plausible depth prediction. We further visualize the absolute relative depth error, which shows that MOVEDepth can significantly reduce the depth error brought by monocular methods.

\subsection{DDAD Results}




\definecolor{Asectioncolor}{RGB}{255, 200, 200}
\definecolor{Bsectioncolor}{RGB}{255, 228, 196}
\definecolor{Csectioncolor}{RGB}{235, 255, 235}
\definecolor{Dsectioncolor}{RGB}{235, 235, 255}

\begin{table}[t]
  \centering
  \resizebox{0.47\textwidth}{!}{
    \begin{tabular}{l|c|c|c|c}
        \arrayrulecolor{black}\hline
          Method & \cellcolor{col1}Abs. Rel. & \cellcolor{col1}Sq. Rel. & \cellcolor{col1}RMSE   & \cellcolor{col2}$\delta < 1.25 $\\
         
        \hline\hline
       \noalign{\smallskip}
          Monodepth2 \cite{ClmentGodard2018DiggingIS} &
         {0.213} &   {4.975} &   {18.051} &   {0.761} \\ 
         \midline
         Packnet \cite{VitorGuizilini20193DPF}& 0.162 & 3.917 & 13.452 & 0.823 \\
        \midline
         GUDA \cite{DBLP:conf/iccv/Guizilini0AG21}& 0.147 & 2.922 & 14.452 & 0.809 \\
        \midline
         ManyDepth \cite{JamieWatson2021TheTO}& 0.145 & 3.246 & 13.982 & 0.821 \\
        \midline
        \noalign{\smallskip}
        \hline
        \noalign{\smallskip}
        
         $\text{MOVEDepth}$ (with Monodepth2) &   0.136  &   3.027  &   12.478  &   0.835  \\
         \midline
         $\textbf{MOVEDepth (with Packnet)}$ & \textbf{0.134}  & \textbf{2.903} & \textbf{12.332} & \textbf{0.837}\\
         \noalign{\smallskip}
        \arrayrulecolor{black}\hline
    \end{tabular}
  } 
  \caption{Depth evaluation results on DDAD validation set \cite{VitorGuizilini20193DPF}. The above methods are evaluated for depth range of 200m without cropping. } 
    \label{tab:ddad} 
\end{table}

DDAD is a novel benchmark for depth estimation, which evaluates depth with longer distance and denser LiDAR points. We conduct experiments on DDAD to verify our method can generalize to more challenging scenarios. As shown in Tab.~\ref{tab:ddad}, MOVEDepth still outperforms previous competitors. Notably, our method relatively improves the state-of-the-art multi-frame method \cite{JamieWatson2021TheTO} by 7.6\% (on Abs. Rel.).

\subsection{Ablation Study}
In this subsection, we conduct an ablation study on the KITTI dataset to analyze the effectiveness of each component. Firstly, we analyze the impact of the monocular priority for MVS, demonstrating that the monocular priority significantly reduces multi-frame matching ambiguity with fewer depth candidates. Then we verify that, compared with cascade sampling \cite{DBLP:conf/cvpr/GuFZDTT20} and confidence sampling \cite{Bae2022}, multi-frame depth learning benefits more from the proposed velocity-guided depth sampling. Lastly, we qualitatively show that the uncertainty-based fusing makes MOVEDepth more robust against challenging artifacts where multi-view geometry fails (e.g., moving objects, textureless areas). In the following ablation study, unless specified, we utilize Monodepth2 \cite{ClmentGodard2018DiggingIS} as the monocular priority, and only two frames (0, -1) are leveraged to construct MVS cost volume.

\subsubsection{Monocular Priority for Multi-Frame Depth Learning}




\definecolor{Asectioncolor}{RGB}{255, 200, 200}
\definecolor{Bsectioncolor}{RGB}{255, 228, 196}
\definecolor{Csectioncolor}{RGB}{235, 255, 235}
\definecolor{Dsectioncolor}{RGB}{235, 235, 255}

\begin{table}[ht]
  \centering
  \resizebox{0.47\textwidth}{!}{
    \begin{tabular}{l|c|c|c|c|c}
        \arrayrulecolor{black}\hline
          Method & \cellcolor{col1}Abs. Rel. & \cellcolor{col1}Sq. Rel. & \cellcolor{col1}RMSE   & \cellcolor{col2}$\delta < 1.25 $&GPU (MB)\\
         
        \hline\hline
       \noalign{\smallskip}
          Mono. baseline  & {0.115} &   {0.903 } &   {4.863} &   {0.877} & \textbf{161} \\ 
          MOVEDepth w/o priority (96 bins) & {0.328} &   {2.892 } &   {12.873} &   {0.633} & {678} \\ 
          MOVEDepth w/o priority (48 bins) & {0.349} &   {3.029 } &   {13.021} &   {0.582} & {483} \\ 
         \noalign{\smallskip}
          \hline
           \noalign{\smallskip}
          MOVEDepth (48 bins) & {0.096} &   {0.792 } &   {4.445} &   {0.900} & {466} \\ 
          MOVEDepth (32 bins) & {0.093} &   {0.746 } &   {4.382} &   {0.902} & {390} \\ 

          $\textbf{MOVEDepth}$ \textbf{(16 bins)} & \textbf{0.092} &   \textbf{0.686 } &   \textbf{4.332} &   \textbf{0.904} & {322} \\ 
          MOVEDepth (8 bins) & {0.094} &   {0.736 } &   {4.419} &   {0.903} & {285} \\ 
          \noalign{\smallskip}
        \arrayrulecolor{black}\hline
        
    \end{tabular}
  } 
  \caption{Ablation analysis on monocular priority for multi-frame depth learning, including depth metrics and inference GPU memory consumption (GPU consumption is measured by \texttt{torch.cuda.max\_memory\_allocated}).} 
    \label{tab:abla_mpm} 
\end{table}

As shown in Fig.~\ref{fig:kitti_vis}, due to matching ambiguity in a large depth range, the multi-view geometry (without monocular priority) fails to learn multi-frame depth in a self-supervised manner. Besides, quantitative results (Tab. \ref{tab:abla_mpm}) also show that the MOVEDepth (without monocular priority) is inferior to the monocular priority \cite{ClmentGodard2018DiggingIS}, yet with $3\times \sim 4\times$ GPU memory consumption. 

When integrated with the monocular depth priority, the geometric uncertainty is significantly reduced, and MOVEDepth can produce superior depth predictions with less depth bin candidates. We empirically select 16 depth bins for MOVEDepth, as it strikes a balance between accuracy and memory consumption, and MOVEDepth with 16 depth bins relatively outperform the monocular baseline by 20\% (on Abs. Rel.).

\subsubsection{Velocity-Guided Depth Sampling}




\definecolor{Asectioncolor}{RGB}{255, 200, 200}
\definecolor{Bsectioncolor}{RGB}{255, 228, 196}
\definecolor{Csectioncolor}{RGB}{235, 255, 235}
\definecolor{Dsectioncolor}{RGB}{235, 235, 255}

\begin{table}[ht]
  \centering
  \resizebox{0.47\textwidth}{!}{
    \begin{tabular}{l|c|c|c|c}
        \arrayrulecolor{black}\hline
          Method & \cellcolor{col1}Abs. Rel. & \cellcolor{col1}Sq. Rel. & \cellcolor{col1}RMSE   & \cellcolor{col2}$\delta < 1.25 $\\
         
        \hline\hline
       \noalign{\smallskip}
       
   $\text{MOVEDepth}^{\dag}$ (frame 0, -1) & {0.101} &   {0.801 } &   {4.474} &   {0.897}  \\ 
  $\text{MOVEDepth}^{\ddag}$ (frame 0, -1) & {0.099} &   {0.773 } &   {4.432} &   {0.898}  \\ 
  $\text{MOVEDepth}^{\dag}$ (frame 0, -1, -2) & {0.100} &   {0.824 } &   {4.489} &   {0.896}  \\ 
  $\text{MOVEDepth}^{\dag}$ (frame 0, -1, +1) & {0.098} &   {0.769 } &   {4.418} &   {0.900}  \\ 
          \noalign{\smallskip}
          \hline
          \noalign{\smallskip}

          $\text{MOVEDepth}$ (cascade) & {0.096} &   {0.762 } &   {4.445} &   {0.899} \\ 
          $\text{MOVEDepth}$ (confidence) & {0.097} &   {0.781 } &   {4.493} &   {0.897} \\ 
          
          \noalign{\smallskip}
          \hline
          \noalign{\smallskip}
          $\text{MOVEDepth}$ (vel. $\beta$=0.1) & {0.094} &   {0.712 } &   {4.398} &   {0.902} \\ 
          $\textbf{MOVEDepth}$ \textbf{(vel. $\mathbf{\beta}$=0.15)} & \textbf{0.092} &   \textbf{0.686 } &   {4.332} &   \textbf{0.904}\\ 
          $\text{MOVEDepth}$ (vel. $\beta$=0.2) & {0.093} &   {0.692 } &   \textbf{4.346} &   {0.902} \\ 
        \noalign{\smallskip}
        \arrayrulecolor{black}\hline
    \end{tabular}
  } 
  \caption{Ablation analysis on velocity-guided depth sampling. $\text{MOVEDepth}^{\dag}$ denotes fixed depth range $[\frac{1}{2}D_{\text{Mono}},\frac{3}{2}D_{\text{Mono}}]$, and $\text{MOVEDepth}^{\ddag}$ denotes fixed depth range $[\frac{3}{4}D_{\text{Mono}},\frac{5}{4}D_{\text{Mono}}]$.} 
    \label{tab:abla_vel} 
\end{table}





\definecolor{Asectioncolor}{RGB}{255, 200, 200}
\definecolor{Bsectioncolor}{RGB}{255, 228, 196}
\definecolor{Csectioncolor}{RGB}{235, 255, 235}
\definecolor{Dsectioncolor}{RGB}{235, 235, 255}

\begin{table}[ht]
  \centering
  \resizebox{0.47\textwidth}{!}{
    \begin{tabular}{l|c|c|c|c}
        \arrayrulecolor{black}\hline
          Method & \cellcolor{col1}Abs. Rel. & \cellcolor{col1}Sq. Rel. & \cellcolor{col1}RMSE   & \cellcolor{col2}$\delta < 1.25 $\\
         
        \hline\hline
       \noalign{\smallskip}
          MOVEDepth (unfused) & {0.094} &   {0.728 } &   {4.362} &   {0.902}  \\ 

          MOVEDepth (fused) & \textbf{0.092} &   \textbf{0.686 } &   \textbf{4.332} &   \textbf{0.904} \\ 
\noalign{\smallskip}
        \arrayrulecolor{black}\hline
        
    \end{tabular}
  } 
  \caption{Ablation analysis on uncertainty-based fusing.} 
    \label{tab:abla_fuse} 
\end{table}

As shown in Tab. \ref{tab:abla_vel}, MOVEDepth with fixed depth sampling range (row 1,2) shows restricted performance, and increasing input frames (row 3,4) can not improve depth accuracy due to the inherent geometric ambiguity (without pose and depth supervision). Recent methods adjust depth range by cascade sampling \cite{DBLP:conf/cvpr/GuFZDTT20} (half the depth range in the next stage) or confidence-based sampling \cite{Bae2022} (adjust depth range according to monocular depth confidence). However, these sampling strategies (row 5,6) bring marginal depth accuracy improvement, as they overlook the \textit{Triangulation Prior}  nearby frames. In contrast, MOVEDepth associates \textit{Triangulation Prior} with the predicted camera velocity, and adaptively adjusts depth range under the guidance of velocity. Significantly, MOVEDepth with velocity guidance ($\beta$=0.15) relatively outperforms its fixed depth range baseline (row 1) by 7.1\% (on Abs. Rel.).

\subsubsection{Uncertainty-Based Depth Fusing}

\begin{figure}[ht]
  \centering
  \includegraphics[width=0.43\textwidth]{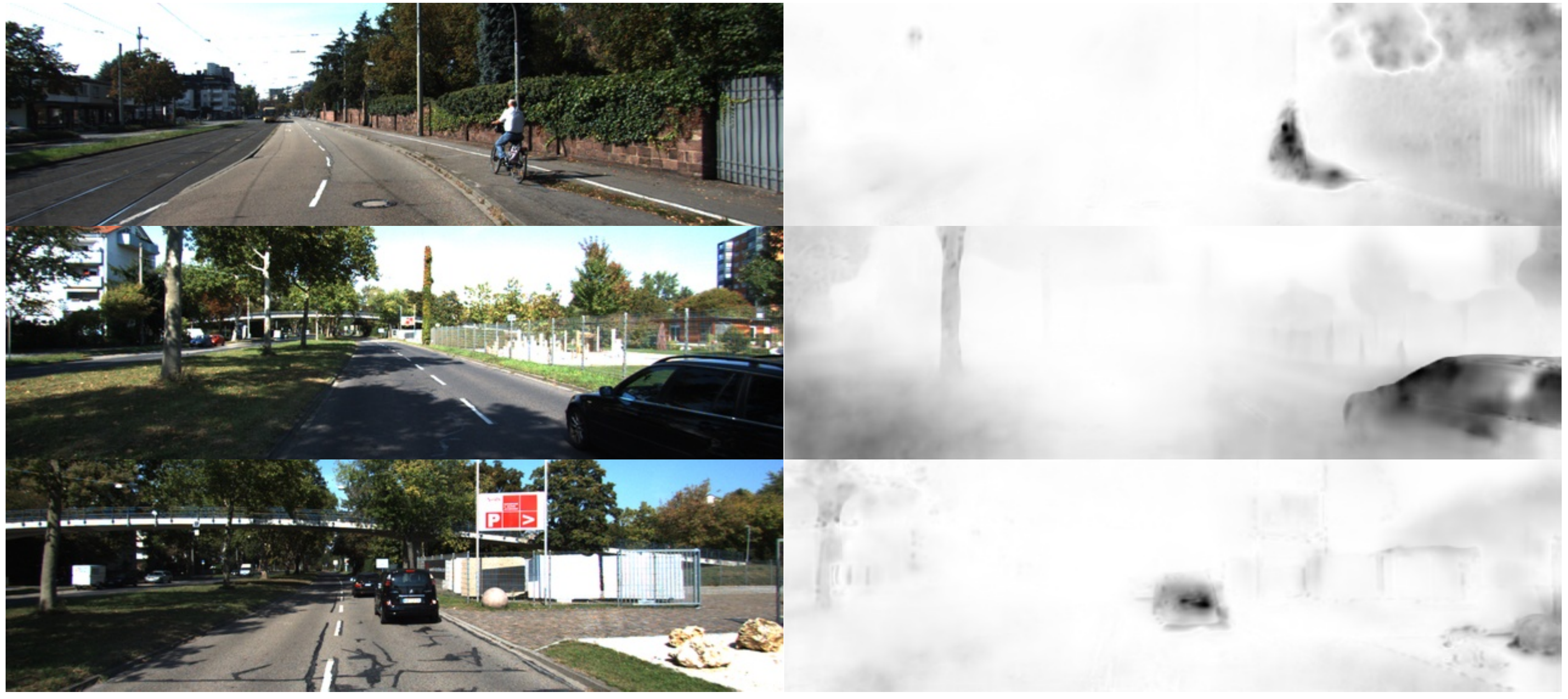}
  \caption{Visualization of the learned uncertainty map, where the left figures are input RGB images, and the right figures are the learned uncertainty map (white: certain, black: uncertain). }
  \label{fig:fuse_vis}
  \vspace{-0.5cm}
  \end{figure}
The uncertainty-based depth fusing strategy is utilized to complement challenging artifacts where multi-view geometry fails. As shown in Fig. \ref{fig:fuse_vis}, the moving objects and textureless areas are learned as highly-uncertain regions. Besides, learning the uncertainty in MVS regularizes the cost volume and further improves the depth accuracy on all the metrics (see Tab. \ref{tab:abla_fuse}).

\section{Conclusion}

In this paper, we propose MOVEDepth, which crafts monocular cues and velocity guidance for improving multi-frame depth learning in a self-supervised manner. The proposed method leverages monocular depth priority to reduce multi-frame matching geometry ambiguity. Specifically, MOVEDepth constructs cost volume with depth range centered at the monocular priority, and the depth range is adaptively adjusted by the predicted camera velocity. Therefore, video frames with insufficient \textit{Triangulation Prior} shrink the depth range to monocular priority, and those with sufficient \textit{Triangulation Prior} expand the depth range to infer more accurate depth. Moreover, we learn depth uncertainty from the entropy of MVS depth probability, which results in a more robust depth against challenging artifacts where multi-view geometry fails. Extensive experiments on KITTI and DDAD show that MOVEDepth achieves stage-of-the-art depth accuracy with significantly reduced depth range. We hope that MOVEDepth can inspire more self-supervised multi-frame depth learning methods in the future.

\bibliography{aaai23}

\newpage
\section{Triangulation Prior Analysis}
Binocular systems rely on matching geometry to predict depth $D$:
\begin{equation}
    D=f\frac{b}{d},
\label{eq:stereo}
\end{equation}
where $f$ is the focal length, $d$ is the disparity of the matching pixels, and $b$ is the stereo baseline. To make a reliable depth estimation, sufficient \textit{Triangulation Prior} \cite{JohannesLSchonberger2016PixelwiseVS} is desired. Arguably, \textit{Triangulation Prior} is represented by baseline $b$ in the stereo systems. In the following, we formulate the stereo baseline in a more general multi-frame setting.

\begin{figure}[ht]
  \centering
  \includegraphics[width=0.2\textwidth]{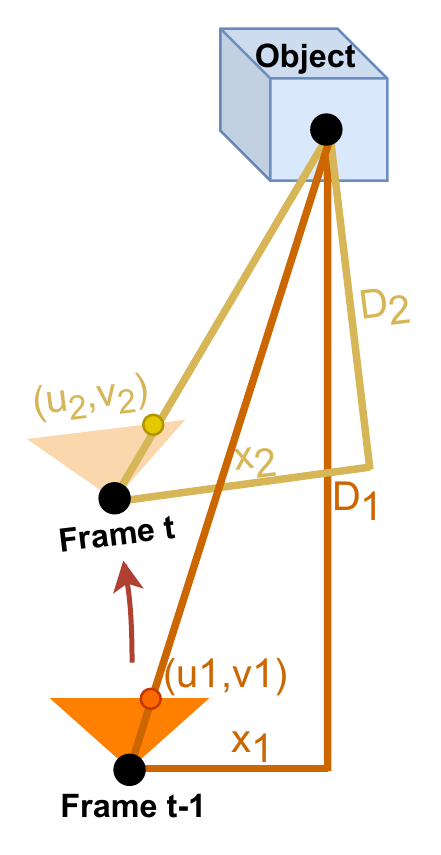}
  \caption{Multi-view geometry in video frames with ego-motion.}
  \label{fig:tri_pri}
  \end{figure}

As shown in Fig \ref{fig:tri_pri}, the moving camera captures images at frame $t-1$ and frame $t$, and the (camera-centered) coordinates of the observed object are $(X_1,Y_1,D_1)^\text{T}$ and $(X_2,Y_2,D_2)^\text{T}$. Given camera extrinsic $\mathbf{R}$ and $\mathbf{T}$, the transformation brought by ego-motion is formulated as:

\begin{equation}
\left(\begin{array}{c}
X_{2} \\
Y_{2} \\
D_{2}
\end{array}\right)=\mathbf{R}\left(\begin{array}{c}
X_{1} \\
Y_{1} \\
D_{1}
\end{array}\right)+\mathbf{T} .
\label{eq:ext}
\end{equation}
Given camera intrinsic:
\begin{equation}
    \mathbf{K}=\left(\begin{array}{ccc}
f & 0 & c_{u} \\
0 & f & c_{v} \\
0 & 0 & 1
\end{array}\right),
\end{equation}
we can calculate the corresponding pixel locations of $(X_1,Y_1,D_1)^\text{T}$ and $(X_2,Y_2,D_2)^\text{T}$:
\begin{equation}
    \left(\begin{array}{c}
u_{2} \\
v_{2} \\
1
\end{array}\right) =\frac{1}{D_{2}}\left(\begin{array}{lll}
f & 0 & c_{u} \\
0 & f & c_{v} \\
0 & 0 & 1
\end{array}\right)\left(\begin{array}{l}
X_{2} \\
Y_{2} \\
D_{2}
\end{array}\right),
\label{eq:int1}
\end{equation}

\begin{equation}
    \left(\begin{array}{c}
u_{1} \\
v_{1} \\
1
\end{array}\right) =\frac{1}{D_{1}}\left(\begin{array}{lll}
f & 0 & c_{u} \\
0 & f & c_{v} \\
0 & 0 & 1
\end{array}\right)\left(\begin{array}{l}
X_{1} \\
Y_{1} \\
D_{1}
\end{array}\right).
\label{eq:int2}
\end{equation}
Constrained by Eq. \ref{eq:ext}, Eq. \ref{eq:int1}, and Eq. \ref{eq:int2}, the depth at frame $t$ can be derived as:
\begin{equation}
    D_{2}=f\frac{\mathbf{R_3}\left(\begin{array}{c}
\frac{u_{1}-c_{u}}{f} \\
\frac{v_{1}-c_{v}}{f} \\
1
\end{array}\right)T_1-\mathbf{R_1}\left(\begin{array}{c}
\frac{u_{1}-c_{u}}{f} \\
\frac{v_{1}-c_{v}}{f} \\
1
\end{array}\right)T_3}{\mathbf{R_3}\left(\begin{array}{c}
\frac{u_{1}-c_{u}}{f} \\
\frac{v_{1}-c_{v}}{f} \\
1
\end{array}\right)\left(u_{2}-c_{u}\right)-\mathbf{R_1}\left(\begin{array}{c}
\frac{u_{1}-c_{u}}{f} \\
\frac{v_{1}-c_{v}}{f} \\
1
\end{array}\right) f},
\end{equation}
where $\mathbf{R_1}$, $\mathbf{R_3}$ are the 1-st and 3-rd row vector of $\mathbf{R}$, and $T_1, T_3$ are the 1-st and 3-rd scalar of $\mathbf{T}$.

\begin{figure}[ht]
  \centering
  \includegraphics[width=0.25\textwidth]{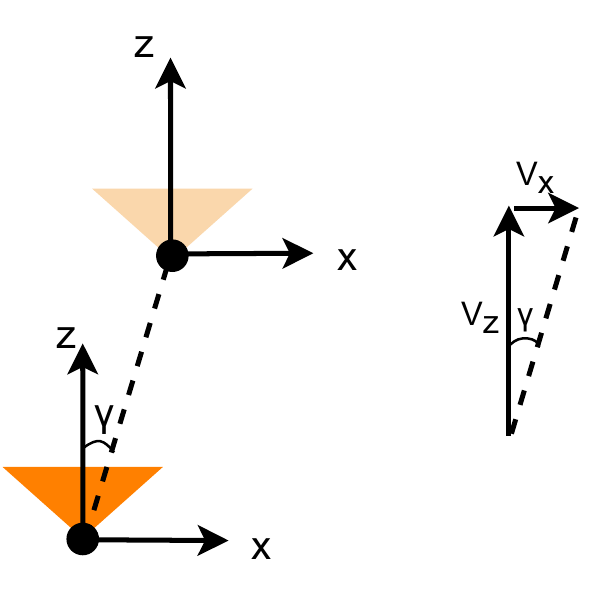}
  \caption{Illustration of the camera moving at a constant speed.}
  \label{fig:mv}
  \end{figure}

Considering cameras in the driving scenario where the car has no ego-rotation, we can set $\mathbf{R}$ to the identity matrix. Then the depth is simplified as follows:
\begin{equation}
    D_2=\frac{f(T_1-\frac{u_1-c_u}{f}T_3)}{u_2-u_1}.
\end{equation}
Arguably, the camera moves at a constant speed during the frame interval. Therefore,  $T_1, T_3$ can be formulated by camera velocity:
\begin{equation}
    T_1=\alpha V_x, \quad T_3=\alpha V_z,
\end{equation}
where $\alpha$ is the camera frame rate. As shown in Fig \ref{fig:mv}, $V_x$ and $V_z$ is the camera speed along the x-axis and z-axis, and they satisfy:
\begin{equation}
    V_x=V_z\tan \gamma,
\end{equation}
where $\gamma$ is the yaw angle of the moving camera and $\gamma \approx 0$ in the driving scenario. Finally, the depth at frame $t$ is:
\begin{equation}
    D_2=f\frac{\alpha (\tan \gamma -\frac{u_1-c_u}{f})V_z}{u_2-u_1}.
\end{equation}
Compared with the binocular system (Eq. \ref{eq:stereo}), we can formulate the general disparity and baseline in the multi-frame system: $u_2-u_1$ is the general disparity, and $\alpha (\tan \gamma -\frac{u_1-c_u}{f})V_z$ is the general baseline, where the per-pixel baseline is positively related to the camera velocity. 

Therefore, it is concluded that a higher camera velocity brings a larger baseline, which provides sufficient \textit{Triangulation Prior} for the multi-frame system. In contrast, slow camera motion provides a little baseline, and the \textit{Triangulation Prior} is limited.

\section{Additional Implementation Details}
Recent learning-based Multi-View Stereo (MVS) methods adopt Deformable Convolutional Networks (DCNs)~\cite{DBLP:conf/iccv/DaiQXLZHW17} or Transformer blocks \cite{DBLP:journals/corr/abs-2111-14600,DBLP:journals/corr/abs-2204-07616} to enhance the feature learning. However, in this paper, the feature extractor $\theta_\text{enc}$ and cost volume decoder $\theta_\text{dec}$ only comprise of  2D/3D Convolutional Neural Networks (CNNs), because the argument of our paper is to verify the effectiveness of \textit{Monocular Depth Priority} and \textit{Velocity-Guided sampling}. 

\subsection{Network Architecture of Feature Extractor} 
We use a four-stage Feature Pyramid Network (FPN)~\cite{DBLP:conf/cvpr/LinDGHHB17} to extract multi-frame features, and the detailed parameters with layer descriptions are summarized in Table~\ref{table:fpn}. Notably, we only leverage features from \textit{Stage 3} as the output, whose resolution is $\frac{1}{4}$ of the original image.

\setlength{\tabcolsep}{10pt}
\begin{table}[ht]
\scriptsize
\begin{center}
\caption{The detailed parameters of $\theta_\text{enc}$, where S denotes stride, and if not specified with *, each convolution layer is followed by a Batch Normalization layer (BN) and a Rectified Linear Unit (ReLU).}
\label{table:fpn}
\resizebox{0.47\textwidth}{!}{
\begin{tabular}{lcc}
\hline\noalign{\smallskip}
Stage Description & Layer Description & Output Size \\
\noalign{\smallskip}
\hline
\noalign{\smallskip}
-& Input Images  & $H\times W\times 3$\\
\noalign{\smallskip}
\hline
\noalign{\smallskip}
  FPN Stage 1 & Conv2D, $3\times 3$, S1, 8  & $H\times W\times 8$\\
  \noalign{\smallskip}
  FPN Stage 1 & Conv2D, $3\times 3$, S1, 8  & $H\times W\times 8$\\
\noalign{\smallskip}
\hline
\noalign{\smallskip}
  FPN Stage 2 & Conv2D, $5\times 5$, S2, 16  & $ H/2 \times W/2 \times 16$\\
  \noalign{\smallskip}
  FPN Stage 2 & Conv2D, $3\times 3$, S1, 16  & $ H/2 \times W/2 \times 16$\\
  \noalign{\smallskip}
  FPN Stage 2 & Conv2D, $3\times 3$, S1, 16  & $ H/2 \times W/2 \times 16$\\
\noalign{\smallskip}
\hline
\noalign{\smallskip}
  FPN Stage 3 & Conv2D, $5\times 5$, S2, 32  & $H/4\times W/4\times 32$\\
  \noalign{\smallskip}
  FPN Stage 3 & Conv2D, $3\times 3$, S1, 32  & $H/4\times W/4\times 32$\\
  \noalign{\smallskip}
  FPN Stage 3 & Conv2D, $3\times 3$, S1, 32  & $H/4\times W/4\times 32$\\
  \noalign{\smallskip}
  FPN Stage 3 Inner Layer* & Conv2D, $1\times 1$, S1, 64  & $H/4\times W/4\times 64$\\
  \noalign{\smallskip}
  \textbf{FPN Stage 3 Output Layer*} & Conv2D, $1\times 1$, S1, 32  & $H/4\times W/4\times 32$\\
\noalign{\smallskip}
\hline
\noalign{\smallskip}
  FPN Stage 4 & Conv2D, $5\times 5$, S2, 64  & $H/8\times W/8\times 64$\\
  \noalign{\smallskip}
  FPN Stage 4 & Conv2D, $3\times 3$, S1, 64  & $H/8\times W/8\times 64$\\
  \noalign{\smallskip}
  FPN Stage 4 & Conv2D, $3\times 3$, S1, 64  & $H/8\times W/8\times 64$\\
  \noalign{\smallskip}
  FPN Stage 4 Inner Layer* & Conv2D, $1\times 1$, S1, 64  & $H/8\times W/8\times 64$\\
 \noalign{\smallskip}
\hline
\end{tabular}}
\end{center}
\end{table}

\subsection{Network Architecture of Cost Volume Decoder} An UNet~\cite{DBLP:conf/miccai/RonnebergerFB15} structured 3D CNN is adopted for decoding MVS cost volume, and the network details are in Table~\ref{table:3D}.

\setlength{\tabcolsep}{5pt}
\begin{table}
\scriptsize
\begin{center}
\caption{The detailed parameters of $\theta_\text{dec}$, where S denotes stride, and if not specified with *, each convolution layer is followed by a Batch Normalization layer (BN) and a Rectified Linear Unit (ReLU).}
\label{table:3D}
\resizebox{0.47\textwidth}{!}{
\begin{tabular}{lcc}
\hline\noalign{\smallskip}
Stage Description & Layer Description & Output Size \\
\noalign{\smallskip}
\hline
\noalign{\smallskip}
-& Input Cost Volume & $H/4\times W/4\times16\times 16$\\
\noalign{\smallskip}
\hline
\noalign{\smallskip}
  UNet Stage 1 & Conv3D, $3\times 3\times3$, S1, 16  & $H/4\times W/4\times16\times 16$\\
  \noalign{\smallskip}
  UNet Stage 1 & Conv3D, $3\times 3\times3$, S2, 32  & $H/8\times W/8\times 16\times 32$\\
  \noalign{\smallskip}
  UNet Stage 1 & Conv3D, $3\times 3\times 3$, S1, 32  & $H/8\times W/8\times 16\times 32$\\
  \noalign{\smallskip}
  UNet Stage 1 Inner Layer & TransposeConv3D, $3\times 3\times3$, S2, 16  & $H/2\times W/2\times 16\times 16$\\
  \noalign{\smallskip}
  UNet Stage 1 Output Layer* & TransposeConv3D, $3\times 3\times 3$, S1, 16  & $H/2\times W/2\times 16\times 16$\\
\noalign{\smallskip}
\hline
\noalign{\smallskip}
  UNet Stage 2 & Conv3D, $3\times 3\times3$, S2, 64  & $ H/16 \times W/16 \times16\times64$\\
  \noalign{\smallskip}
  UNet Stage 2 & Conv3D, $3\times 3\times3$, S1, 64  & $ H/16 \times W/16 \times16\times64$\\
  \noalign{\smallskip}
  UNet Stage 2 Inner Layer & TransposeConv3D, $3\times 3\times3$, S2, 32  & $H/8\times W/8\times 16\times 32$\\
\noalign{\smallskip}
\hline
\noalign{\smallskip}
  UNet Stage 3 & Conv3D, $3\times 3\times3$, S2, 128  & $ H/32 \times W/32 \times16\times128$\\
  \noalign{\smallskip}
  UNet Stage 3 & Conv3D, $3\times 3\times3$, S1, 128  & $ H/32 \times W/32 \times16\times128$\\
  \noalign{\smallskip}
  UNet Stage 3 Inner Layer & TransposeConv3D, $3\times 3\times3$, S2, 64  & $H/16\times W/16\times 16\times 64$\\
\noalign{\smallskip}
\hline
\end{tabular}}
\end{center}
\end{table}

\subsection{Evaluation Metrics}
Following previous depth estimation methods \cite{ClmentGodard2018DiggingIS,JamieWatson2021TheTO,ZiyueFeng2022DisentanglingOM}, we use Absolute Relative Error (Abs. Rel.), Squared Relative Error (Sq. Rel.), Root Mean Squared Error (RMSE), Root Mean Squared Log Error ($\text{RMSE}_{\text{log}}$ ), and $\delta_1,\delta_2,\delta_3$ as the metrics to evaluate the depth prediction performance. These metrics are formulated as:

\begin{align}
\nonumber
    \mathrm{Abs. Rel.} &= \frac{1}{n} \sum_{i} \frac{|p_i - g_i|}{g_i} \\
    \nonumber
    \mathrm{Sq. Rel.} &=  \frac{1}{n} \sum_{i} \frac{(p_i - g_i)^2}{g_i} \\
    \nonumber
    \mathrm{RMSE} &= \sqrt{\frac{1}{n} \sum_{i}(p_i - g_i)^2}\\
    \nonumber
    \mathrm{RMSE_{log}} &=  \sqrt{\frac{1}{n} \sum_{i}(\log p_i - \log  g_i)^2},
\end{align}
and $
  \delta_1, \delta_2, \delta_3= \%\ of\ thresh < 1.25, 1.25^2, 1.25^3,
$
where $g$ and $p$ are the depth values of ground truth and prediction in meters, $thresh=\max(\frac{g}{p},\frac{p}{g})$.

\end{document}